\documentclass[10pt,twocolumn,letterpaper]{article}

\usepackage{wacv}
\usepackage{times}
\usepackage{epsfig}
\usepackage{graphicx}
\usepackage{amsmath}
\usepackage{amssymb}

\usepackage{placeins}

\usepackage{censor}
\usepackage{mathabx}

\usepackage{subcaption}
\usepackage{color,soul}
\usepackage[table,xcdraw]{xcolor}

\def\wacvPaperID{0001} 

\wacvapplicationstrack 

\wacvfinalcopy 

\ifwacvfinal
\usepackage[breaklinks=true,bookmarks=false]{hyperref}
\else
\usepackage[pagebackref=true,breaklinks=true,colorlinks,bookmarks=false]{hyperref}
\fi

\newcommand*{\affaddrMC}[1]{#1} 
\newcommand*{\affmarkMC}[1][*]{\textsuperscript{#1}}

\begin{document}


\title{CIMGEN: \underline{C}ontrolled \underline{I}mage \underline{M}anipulation by Finetuning Pretrained \underline{Gen}erative Models on Limited Data}



\author{%
Chandrakanth Gudavalli\affmarkMC[1], Erik Rosten\affmarkMC[1], Lakshmanan Nataraj \affmarkMC[1], Shivkumar Chandrasekaran \affmarkMC[1, 2], and \\~~~B. S. Manjunath\affmarkMC[1,2]\\
\\
\affaddrMC{\affmarkMC[1]Mayachitra, Inc.}\\
\affaddrMC{\affmarkMC[2]Electrical and Computer Engineering Department, UC Santa Barbara}\\
\affaddrMC{Santa Barbara, California, USA}%
}

\maketitle


\begin{abstract}
\vspace{-0.4cm}

Content creation and image editing can benefit from flexible user controls. A common intermediate representation for conditional image generation is a semantic map, that has information of objects present in the image. When compared to raw RGB pixels, the modification of semantic map is much easier. One can take a semantic map and easily modify the map to selectively insert, remove, or replace objects in the map. The method proposed in this paper takes in the modified semantic map and alter the original image in accordance to the modified map. The method leverages traditional pre-trained image-to-image translation GANs, such as CycleGAN or Pix2Pix GAN, that are fine-tuned on a limited dataset of reference images associated with the semantic maps. We discuss the qualitative and quantitative performance of our technique to illustrate its capacity and possible applications in the fields of image forgery and image editing. We also demonstrate the effectiveness of the proposed image forgery technique in thwarting the numerous deep learning-based image forensic techniques, highlighting the urgent need to develop robust and generalizable image forensic tools in the fight against the spread of fake media.

\end{abstract}

\vspace{-0.8cm}
\section{Introduction}
\label{sec:intro}
\vspace{-0.3cm}
\begin{figure}[t]
    \centering
    \includegraphics[width=0.47\textwidth]{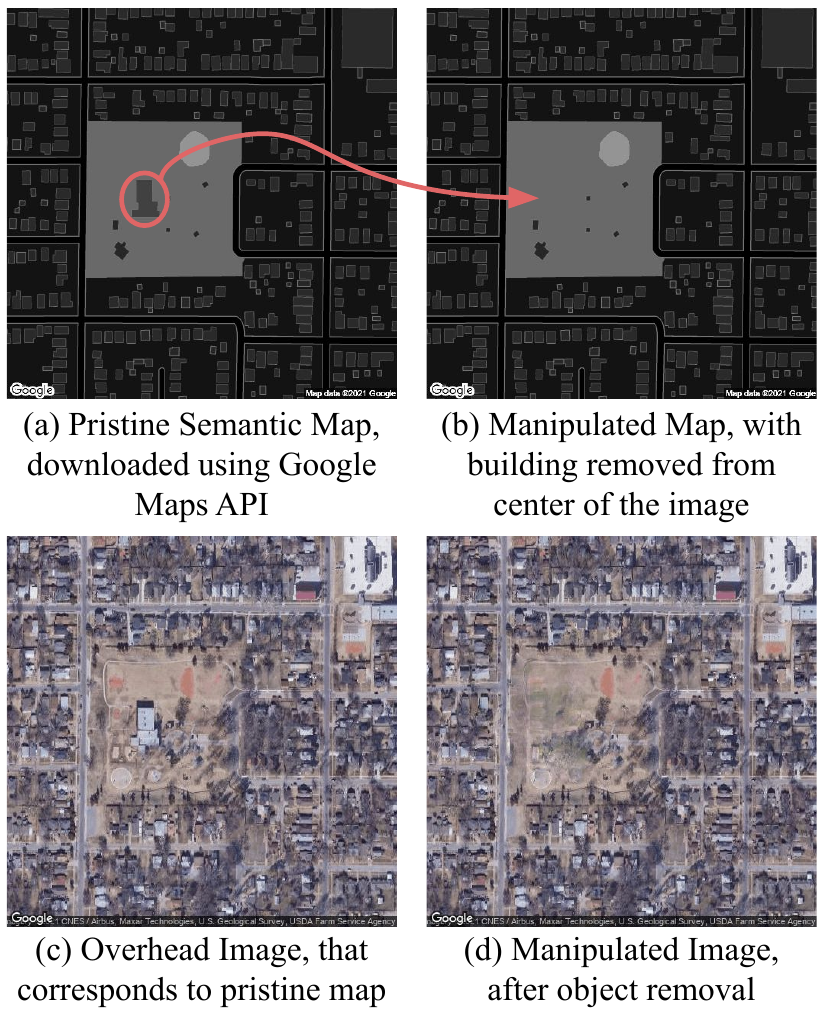}
    \vspace{-0.3cm}
    \caption{Illustration of our proposed method, in which a building is removed from the center of an image.}
    \label{fig:manip_walkthrough_step1-page1-eg}
    \vspace{-0.5cm}
\end{figure}

In recent years, Generative Models (GMs) have made significant advancements in their ability to generate high-quality synthetic images and videos~\cite{dhariwal2021diffusion}. In the area of computer vision, these models can be applied in a variety of ways, such as generating images from text prompts~\cite{ramesh2021zero} and performing tasks such as image to image translations. Generative Adversarial Networks (GANs) are a class of GMs that are originally introduced in 2014 \cite{goodfellow2014generative}.
One area that has seen rapid growth in recent years is GAN based image-to-image translation tasks, some of which include creating art~\cite{li2021collaging, xue2022giraffe, fruhstuck2022insetgan}, inpainting images~\cite{yu2018generativeinpainting, Wang_2022_CVPR, Liu_2021_CVPR} and super-resolution~\cite{LedigTHCATTWS16, Michelini_2022_WACV, emad2022moesr}. However, generative models can also be used for more malicious intentions, producing fake images and videos. Deepfakes refer to a particular application of these types of frameworks - manipulating or synthesizing fake human faces. While the most popular, deepfakes have been rapidly expanding into other domains, one study developed a method to inject or remove cancer in 3D medical scans using GANs, manipulating the scans so convincing that they fooled three radiologists as well as a trained lung cancer screening neural network model that won a \$1 million Kaggle competition \cite{236284}.



In the satellite image domain, a recent paper~\cite{zhao2021deep} explored ``deepfake geography", receiving several mentions throughout the press~\cite{deepfake-geo-2, deepfake-geo-3, RijulGuptaGithub}. This work utilized CycleGAN~\cite{cyclegan} to transfer styles of different cities. For example, a satellite image from Seattle may be stylized to have similar landscape features as a typical satellite image from Beijing. In this paper, we present an extended, generalizable GAN based methodology that can be used for object insertion or object removal from an image (as shown in Figure~\ref{fig:manip_walkthrough_step1-page1-eg}). Our method specifically modifies (insertion/removal) specific localized regions in a semantic map, which we then \emph{translate} to produce a fake image that agrees with the modified semantic map and resembles the original image in the unaltered regions. This generated image is then blended with the pristine image in the same localized region that the input map was manipulated, producing a doctored output image whose pixels are identical to the pristine image in the unaltered regions. These types of manipulations find potential use in applications such as defense, agricultural sector and urban planning. For example, in urban planning, it is quite common to have urban blueprints as maps which can undergo numerous changes in the planning stage depending on required design choices, such as swapping the locations of a park and a building to comply with local zoning ordinances. Using the methodology presented in this paper, these changes can be easily rendered into realistic scenes. In agriculture, the map layout of an agricultural field can be modified to make way for new crops, and corresponding visualizations can be previewed in advance. However, the same technique can be used for malicious purposes, such as removing important structures or landmarks from aerial photographs. Therefore, it is essential to have tools for identifying such altered images. In this paper, we also discuss the limitations of a variety of deep leaning based image forensic techniques, so that the doctored images created with our proposed method cannot be flagged.
\vspace{-0.5cm}
\paragraph{Main contributions.} (1) A novel, simple, and effective way to edit or manipulate images. (2) Show the limits of current image forensic techniques so that fake images made with the proposed method cannot be caught.

\vspace{-0.3cm}
\section{Background}
\label{sec:background}
\vspace{-0.1cm}

\subsection{Generative Adversarial Networks}
\label{ssec:gan_architectures}
\vspace{-0.1cm}
The GAN framework, as briefly mentioned in Section~\ref{sec:intro}, consists of two neural networks that are jointly trained: a generative model $G$ and a discriminative model $D$. In the most simple setup, the objective of this GAN is to generate images that are visually similar to those in the training data distribution $X$. However, in image-to-image translation tasks, the objective of the GAN is to learn a mapping between source domain $X$ to the target domain $Y$ such that $G(X)$ is indistinguishable from $Y$. Some example tasks include translating images from day to night, black and white to color, or map to satellite. It can be seen that $D$ and $G$ generally play a zero-sum game where the generator $G$ is trying to synthesize realistic samples to fool the discriminator $D$. In the following subsections, we briefly cover two GAN frameworks tested with our methodology - CycleGAN~\cite{cyclegan} and Pix2pixHD~\cite{DBLP:journals/corr/abs-1711-11585}, while noting that any architecture that can perform image-to-image translation is applicable within our proposed framework.

\vspace{-0.4cm}
\subsubsection{CycleGAN}
\label{sssec:cycle_gan}
\begin{figure}[!htbp]
\begin{center}
\includegraphics[width=0.4\linewidth, keepaspectratio]{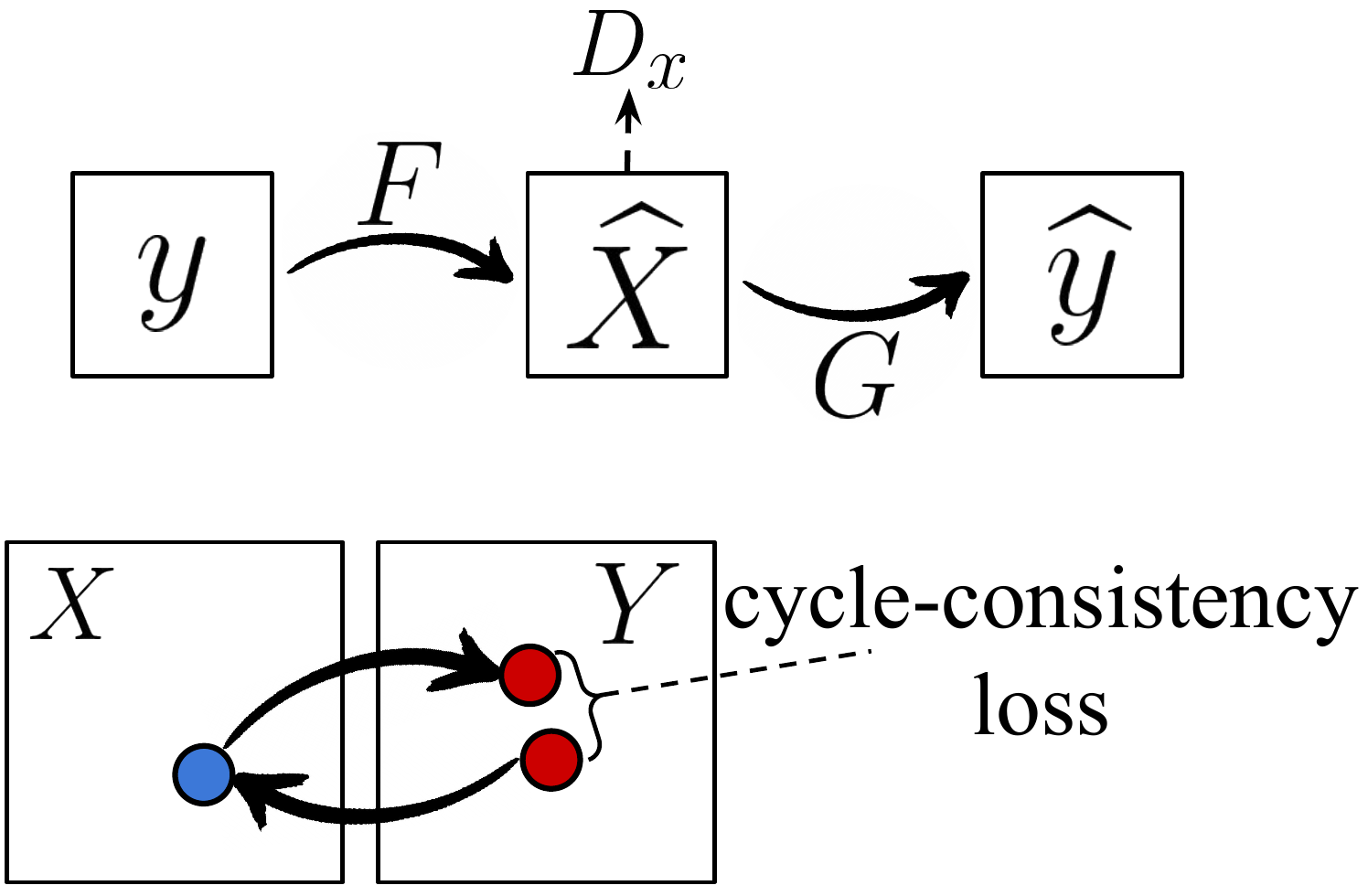}
\end{center}
\vspace{-0.5cm}
\caption{CycleGAN Framework}
\label{fig:cyclegan_consistency_loss}
\vspace{-0.2cm}
\end{figure}

\vspace{-0.2cm}
A core feature of CycleGAN \cite{cyclegan} is its ability to learn image to image translations without paired examples. The main innovation that allows unpaired image to image translation is the addition of a \textit{cycle consistency loss} to the objective function that is being optimized. When training a CycleGAN to learn translations between images $x$ and $y$, two sets of GAN networks are trained. The first network is composed of a generator $G$ that learns a mapping from $x \rightarrow y$, and a discriminator $D_{y}$. The second is composed of a generator $F$ that learns a mapping $y \rightarrow x$ and a discriminator $D_{x}$.
Then, the cycle consistency loss is a measure that tries to enforce that the image translation cycle should be able to bring $x$ back to the original image after passing it through both generators, i.e. $x \rightarrow G(x) \rightarrow F(G(x)) \approx x$. This is called forward cycle consistency. Similarly, backward cycle consistency ensures $y \rightarrow F(y) \rightarrow G(F(y)) \approx y$. We show a visual overview of this formulation for generator $F$ in Figure \ref{fig:cyclegan_consistency_loss}. Mathematically, this cycle consistency term can be represented as shown in Eq~\ref{eqn:cycle_consistency_term}

\vspace{-0.5cm}
\begin{align}
\begin{split}
    \mathcal{L}_{cycle}(G,F) = E_{x \sim p_{data}(x)}[|| F(G(x)) - x||_{1}] + \\
      E_{y \sim p_{data}(y)}[||G(F(y)) - y||_{1}]
  \label{eqn:cycle_consistency_term}
\end{split}
\end{align}

\vspace{-0.5cm}

This cycle consistency loss is added along with two unconditional adversarial loss terms, one for each generator $G$ and $F$. Unconditional in this context means that the input $y$ is not given to the generator or discriminator. This setup ensures that for a given image $x$, we do not need a corresponding output $y$ to optimize the network. The full objective with a tuning parameter $\lambda$ is shown in Eq~\ref{eqn:cycle_full_objective},~\ref{eqn:gan_loss}.

\vspace{-0.5cm}
\begin{align}
\begin{split}
    \mathcal{L}(G,F,D_{x}, D_{y}) = \mathcal{L}_{GAN}(G, D_{y}, x, y) + \\
      \mathcal{L}_{GAN}(F, D_{x}, y, x) + \\
      \lambda~\mathcal{L}_{cycle}(G,F)
  \label{eqn:cycle_full_objective}
\end{split}
\end{align}

\vspace{-0.6cm}
\noindent where,

\vspace{-0.4cm}

\begin{align}
\begin{split}
    \mathcal{L}_{GAN}(G,D, X, Y) = E_{y \sim p_{data}(y)}[log D_{Y}(y)] + \\
      E_{x \sim p_{data}(x)}[log (1 - D_{Y}(y))]
  \label{eqn:gan_loss}
\end{split}
\end{align}

To better enforce color consistent generated images, CycleGAN also introduces an optional $L_{1}$ identity loss that encourages the generated images from each generator network match the input.

\vspace{-0.5cm}
\begin{align}
\begin{split}
    \mathcal{L}_{identity}(G,F) =  E_{x \sim p_{data}(x)}[|| F(x) - x||_{1}] +\\
    E_{y \sim p_{data}(y)}[||G(y) - y||_{1}]
  \label{eqn:l1_identity_loss}
\end{split}
\end{align}

In terms of network architecture, both discriminator networks take the form of a PatchGAN as in the Pix2Pix framework ~\cite{DBLP:journals/corr/IsolaZZE16}. This PatchGAN discriminator classifies each $N$ x $N$ patch in an image as real or fake. Once each patch is classified, all the responses are averaged to provide a final classification from $D$. This patch based network network is faster to run than a full sized image classifier and is argued to take the form of a texture/style loss as it assumes that pixels separated by more than a patch diameter are independent. The generator networks are of an encoder-decoder form composed of ResNet blocks in between downsampling and upsampling layers.

\vspace{-0.5cm}
\subsubsection{Pix2pixHD}
\vspace{-0.3cm}
The second architecture explored is Pix2pixHD , which makes several improvements over pix2pix to improve the quality of generated images that are higher resolution. The first improvement is using multi-scale discriminators and generators to ensure scene consistency at different resolution levels. The second improvement is an improved adversarial loss that incorporates a feature matching loss based on the discriminator.

In pix2pixHD, the generator $G$ is composed of two subnetworks, $G_{1}$ and $G_{2}$ where $G_{1}$ is called a global generator network and $G_{2}$ is called a local enhancer network. Both of these networks are composed of a convolutional front-end, a set of residual blocks and transposed convolutional back-end, where the output of the global generator is half the resolution of the input image in both dimensions and the local enhancer network outputs the original input image size using the output of the global generator. During training, the global generator and local enhancers are each trained individually before being jointly trained. 

The discriminator network $D$ is also designed in a multi-scale fashion, with three discriminators $D_{1}$, $D_{2}$ and $D_{3}$ having the same network structure (PatchGAN), but $D_{2}$ and $D_{3}$ operating on 2x and 4x downsampled real and synthesized images. Then, the learning problem becomes,

\vspace{-0.3cm}
\begin{equation}
    \min_{G} \max_{D_{1}, D_{2}, D_{3}} \sum_{k=1,2,3} \mathcal{L}_{GAN} (G, D_{k})
\end{equation}
\vspace{-0.3cm}

The second improvement in pix2pixHD is the addition of a feature matching loss. The idea here is to extract features from multiple layers in the discriminator as intermediate representations and ensure that these match for real and synthesized images. If the $i$th-layer feature extractor of discriminator $D_{k}$ is $D_{k}^{(i)}$, then the feature matching loss $\mathcal{L}_{FM}(G, D_{k})$ is 

\vspace{-0.5cm}
\begin{equation}
\begin{split}
    \mathcal{L}_{FM}(G, D_{k}) = E_{\boldsymbol{x}, \boldsymbol{y}} \sum_{i=1}^{T} \frac{1}{N_{i}} \left[ || D_{k}^{(i)}(\boldsymbol{x}, \boldsymbol{y}) - \right. \\ \left. D_{k}^{(i)}(\boldsymbol{x}, G(\boldsymbol{x})) ||_{1}\right]
\end{split}
\end{equation}
\vspace{-0.3cm}

\noindent where $T$ is the total number of layers and $N_{i}$ is the number of elements in each layer. Then, the final full objective of the pix2pixHD GAN is

\vspace{-0.6cm}
\begin{equation}
\begin{split}
    \min_{G} \left( \left( \max_{D_{1}, D_{2}, D_{3}} \sum_{k=1,2,3} \mathcal{L}_{cGAN} (G, D_{k}) \right) +  \right. \\ \left. \lambda \sum_{k=1,2,3} \mathcal{L}_{FM}(G, D_{k}) \right)
\end{split}
\end{equation}
\vspace{-0.3cm}

\vspace{-0.4cm}
\section{Proposed Image Manipulation Framework}
\label{sec:prop_method}

\begin{figure*}[!ht]
    \centering
    \includegraphics[width=0.99\textwidth]{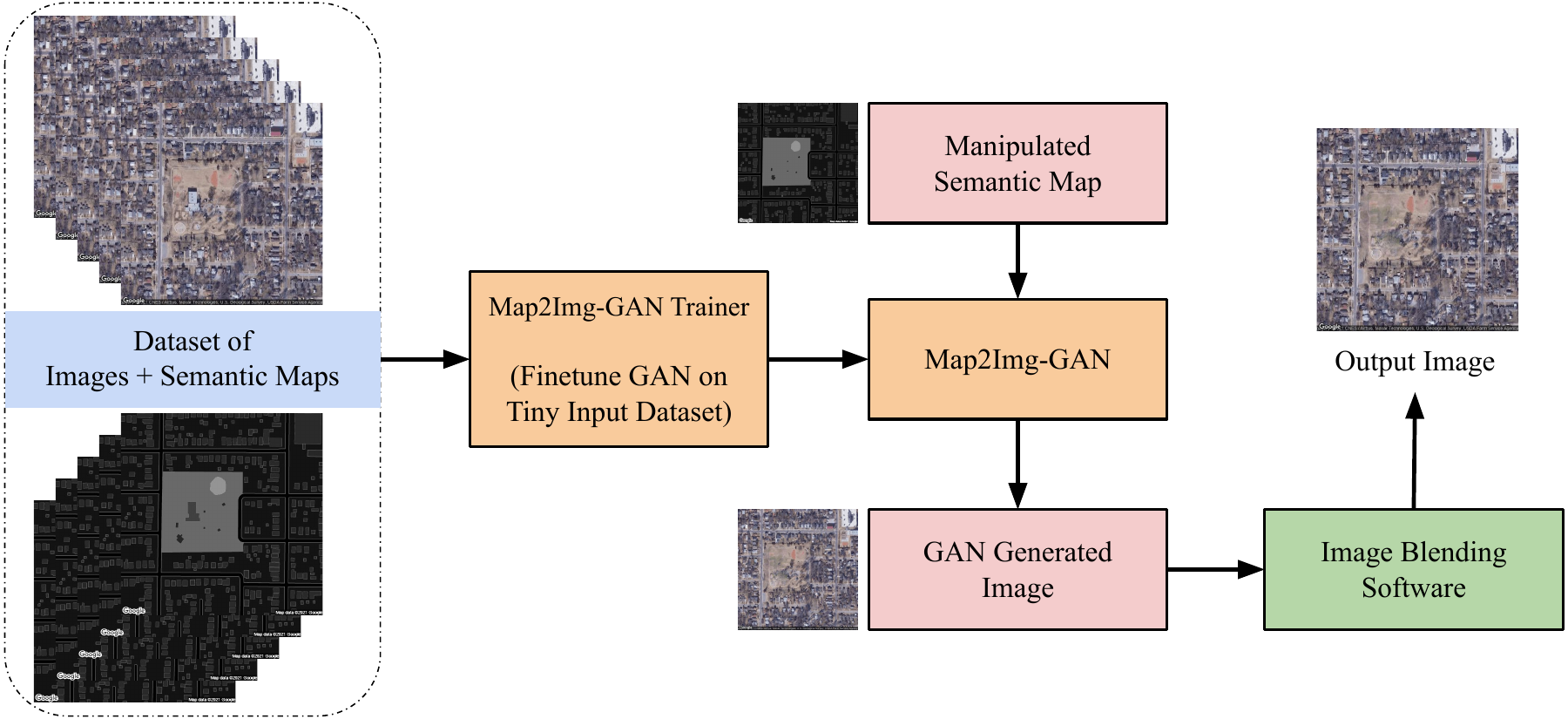}
    \caption{Overview of proposed image manipulation framework.}
    \label{fig:overview}
    \vspace{-0.4cm}
\end{figure*}

We propose a simple yet effective technique for removing/replacing/inserting objects in images. We postulate using a generative model (with CycleGAN or Pix2PiXHD GAN like architecture, pre-trained on standard large datasets in such a way that it is capable of generating images based on their semantic masks), along with a small collection of images tagged to their semantic masks. We refer to this dataset as smalldata, $D$, from now on. Now, we fine tune the pretrained GAN, $G$, on the smalldata, $D$, and finetune the model such that GAN generates an image that closely resembles the original image from the semantic mask that it was trained on. In other words, we mandate the GAN to memorize the correspondence between the semantic map and image. Once the GAN is trained, we generate the tampered semantic mask, $T$, for instance, by removing an object from the original mask, $M$, which is a sample in the smalldata, $D$. Tampering the semantic map can be easily accomplished using open-source software such as Photoshop or GIMP. Using this tampered map, $T$, GAN generates the image that corresponds to that tampered map, which in this case involves the generation of the image similar to the original image, that was paired with $M$, but with the object removed. In a similar way, one can easily insert/remove objects from the images by altering the semantic mask accordingly. This technique has produced qualitatively and quantitatively enticing outcomes, which will be discussed in detail in Section~\ref{sec:experiments}.

The overview of the entire framework is shown in Figure~\ref{fig:overview}, where we feed a manipulated semantic map as an input to the trained generator model. For example, in the context of aerial imagery, one can remove a lake and replace it with a set of buildings. Some applications/examples of such manipulations are discussed in detail in Section~\ref{sec:experiments}. Although the direct output of the generator can be used as the final result, we blend the GAN generated satellite image with the original to bolster the final manipulation's authenticity and ensure that the original pixels are preserved outside of the manipulated region. Possible enhancements and further research opportunities are explored in Section~\ref{sec:conclusion}.

\section{Image Forgery Experiments}
\label{sec:experiments}
\vspace{-0.2cm}
\subsection{Map2Sat Data Curation}
\label{sec:map2sat-urban_data_curation}
\vspace{-0.2cm}
In order to curate a dataset of map to satellite imagery, we gather around 555 pairs of 512 x 512 map and satellite images of US capital cities, scraped from the google maps API. The latitude/longitude coordinates of each capital city are randomly perturbed 10 times within a 5 mile radius to obtain multiple images for each city, which differ in appearance. These resulting image pairs are manually inspected for outliers in order to constrain the domain of the dataset to urban areas. Removing these outliers brings the total number of image pairs to 470, where images were tagged as outliers for three main reasons:

\begin{itemize}
    \vspace{-0.1cm}
    \item Near duplicates due to random perturbation
    \vspace{-0.1cm}
    \item Large regions of the image are not urban 
    \vspace{-0.1cm}
    \item Had visual artifacts from Google stitching images together
\end{itemize}

We denote this dataset throughout the rest of the document as map2sat-urban and show a sample image pair in Figure~\ref{fig:manip_walkthrough_step1-page1-eg}a and Figure~\ref{fig:manip_walkthrough_step1-page1-eg}c.


\subsection{Qualitative Results}

\begin{figure}[!htbp]
\begin{center}
\begin{subfigure}[t]{0.45\linewidth}
\includegraphics[width=\linewidth, keepaspectratio]{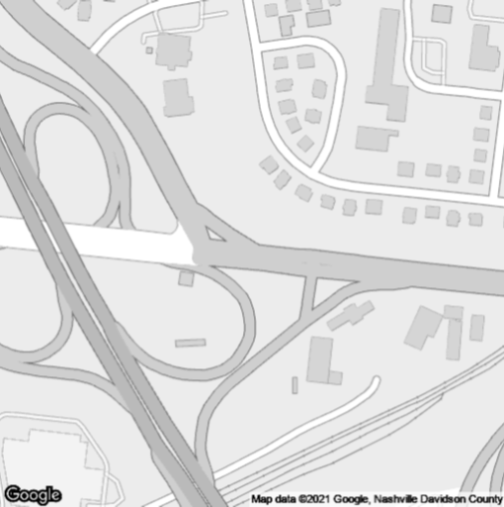}
\caption{Pristine Roadmap, downloaded using Google Maps API}
\end{subfigure}
\hspace{0.1cm}
\begin{subfigure}[t]{0.45\linewidth}
\includegraphics[width=\linewidth, keepaspectratio]{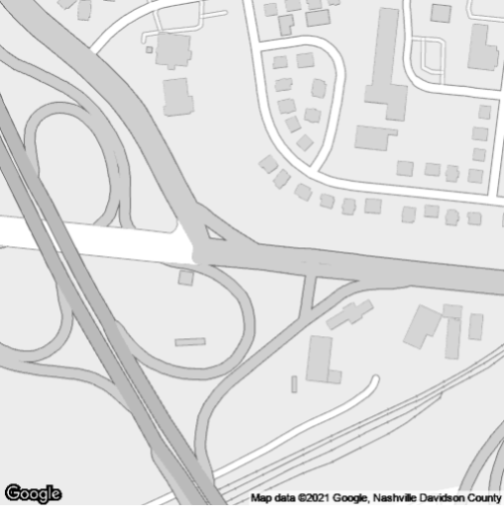}
\caption{Manipulated Roadmap (Objects at bottom left removed)}
\end{subfigure}
\\
\vspace{0.1cm}
\begin{subfigure}[t]{0.45\linewidth}
\includegraphics[width=\linewidth, keepaspectratio]{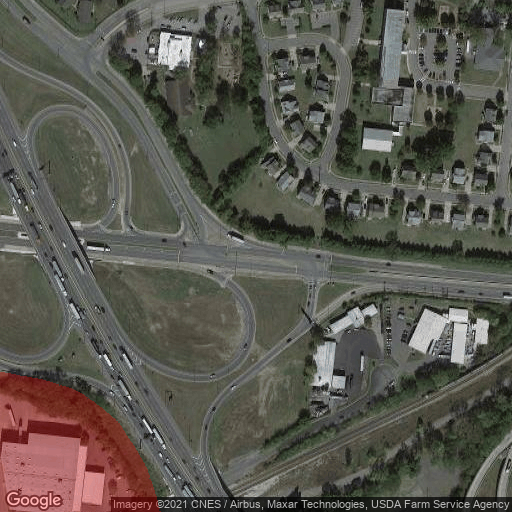}
\caption{Pristine Satellite Image, overlaid on removal mask}
\end{subfigure}
\hspace{0.1cm}
\begin{subfigure}[t]{0.45\linewidth}
\includegraphics[width=\linewidth, keepaspectratio]{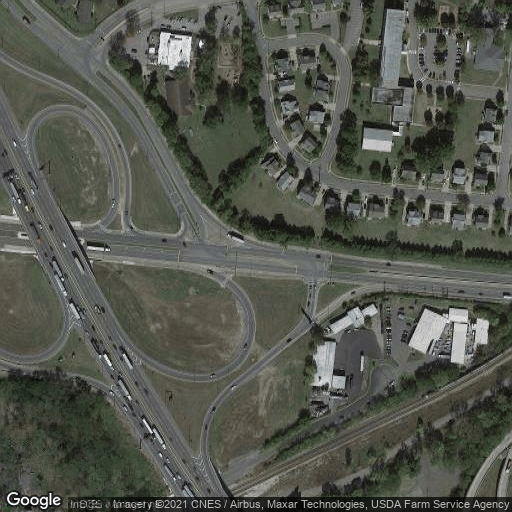}
\caption{Blended Image (Objects at bottom left removed)}
\end{subfigure}
\end{center}
\vspace{-0.5cm}
\caption{An illustration of object removal, using the proposed method with pix2pixHD GAN, from a satellite image of Nashville (the capital of the U.S. state of Tennessee).}
\vspace{-0.3cm}
\label{fig:manip_ex_1}
\end{figure}

\begin{figure}[!htbp]
\begin{center}
\begin{subfigure}[t]{0.45\linewidth}
\includegraphics[width=\linewidth, keepaspectratio]{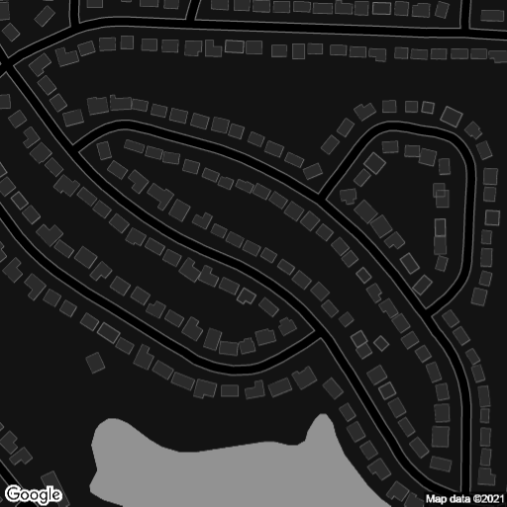}
\caption{Pristine Roadmap, downloaded using Google Maps API}
\end{subfigure}
\hspace{0.1cm}
\begin{subfigure}[t]{0.45\linewidth}
\includegraphics[width=\linewidth, keepaspectratio]{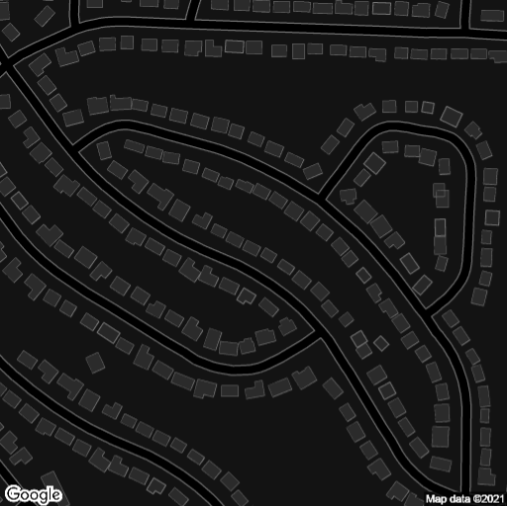}
\caption{Manipulated Roadmap, a water body is replaced by road and buildings}
\end{subfigure}
\\
\vspace{0.1cm}
\begin{subfigure}[t]{0.45\linewidth}
\includegraphics[width=\linewidth, keepaspectratio]{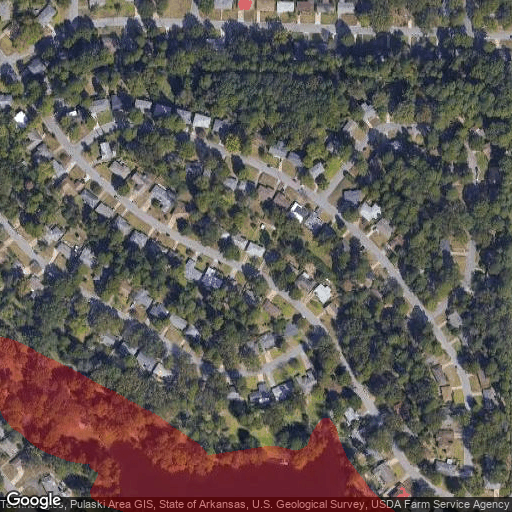}
\caption{Pristine Overhead Image, overlaid on insertion mask}
\end{subfigure}
\hspace{0.1cm}
\begin{subfigure}[t]{0.45\linewidth}
\includegraphics[width=\linewidth, keepaspectratio]{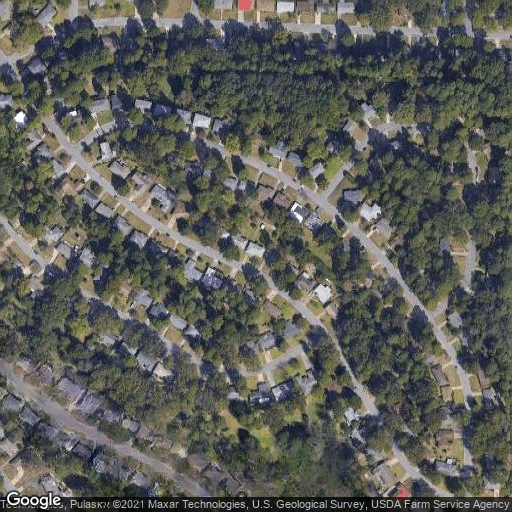}
\caption{Manipulated Image (post blending)}
\end{subfigure}
\end{center}
\vspace{-0.5cm}
\caption{An illustration of object insertion, using the proposed method with pix2pixHD GAN, from a satellite image of Little Rock (the capital of the U.S. state of Arkansas).}
\vspace{-0.3cm}
\label{fig:manip_ex_2}
\end{figure}

\begin{figure}[!htbp]
\begin{center}
\begin{subfigure}[t]{0.45\linewidth}
\includegraphics[width=\linewidth, keepaspectratio]{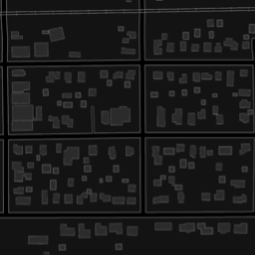}
\caption{Pristine Roadmap}
\end{subfigure}
\hspace{0.1cm}
\begin{subfigure}[t]{0.45\linewidth}
\includegraphics[width=\linewidth, keepaspectratio]{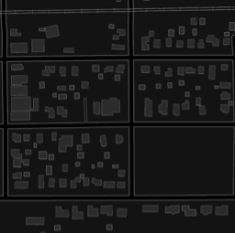}
\caption{Manipulated Roadmap}
\end{subfigure}
\\
\vspace{0.1cm}
\begin{subfigure}[t]{0.45\linewidth}
\includegraphics[width=\linewidth, keepaspectratio]{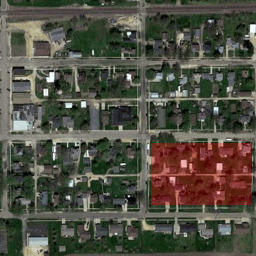}
\caption{Pristine Overhead Image, overlaid on removal mask}
\end{subfigure}
\hspace{0.1cm}
\begin{subfigure}[t]{0.45\linewidth}
\includegraphics[width=\linewidth, keepaspectratio]{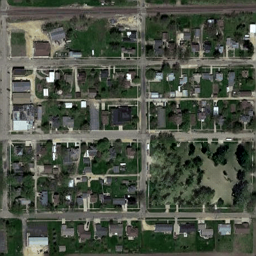}
\caption{Manipulated Image (post blending)}
\end{subfigure}
\end{center}
\vspace{-0.5cm}
\caption{An illustration of the proposed method with CycleGAN, where a large cluster of buildings are removed.}
\vspace{-0.3cm}
\label{fig:cycle_manip_ex_3}
\end{figure}

\begin{figure}[!htbp]
\begin{center}
\begin{subfigure}[t]{0.45\linewidth}
\includegraphics[width=\linewidth, keepaspectratio]{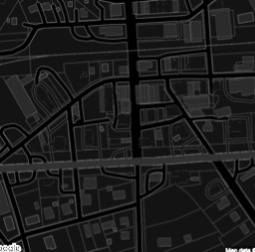}
\caption{Pristine Roadmap}
\end{subfigure}
\hspace{0.1cm}
\begin{subfigure}[t]{0.45\linewidth}
\includegraphics[width=\linewidth, keepaspectratio]{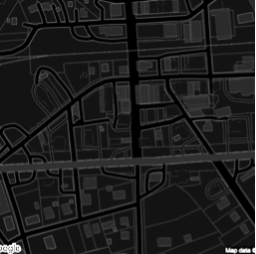}
\caption{Manipulated Roadmap}
\end{subfigure}
\\
\vspace{0.1cm}
\begin{subfigure}[t]{0.45\linewidth}
\includegraphics[width=\linewidth, keepaspectratio]{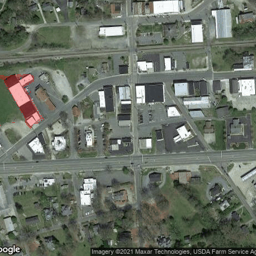}
\caption{Pristine Overhead Image, overlaid on removal mask}
\end{subfigure}
\hspace{0.1cm}
\begin{subfigure}[t]{0.45\linewidth}
\includegraphics[width=\linewidth, keepaspectratio]{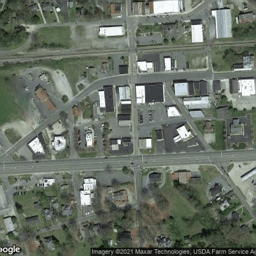}
\caption{Manipulated Image (post blending)}
\end{subfigure}
\end{center}
\vspace{-0.7cm}
\caption{An illustration of the proposed method with CycleGAN, where a building has been removed.}
\label{fig:cycle_manip_ex_5}
\vspace{-0.3cm}
\end{figure}

In Figure \ref{fig:manip_ex_1}, we show an object removal example where we've removed a building region from the bottom left corner of the image. In Figure~\ref{fig:manip_ex_2}, we show an object insertion example in which a body of water is removed, and a road and buildings are inserted in its place. 
Both of these manipulations were generated using the method outlined in Section~\ref{sec:prop_method} with pix2pixHD. In Figures \ref{fig:cycle_manip_ex_3} and \ref{fig:cycle_manip_ex_5} we show manipulations generated using a trained CycleGAN model where a large cluster of buildings has been removed in the first example (Fig~\ref{fig:cycle_manip_ex_3}), and a larger building is removed in the third (Fig~\ref{fig:cycle_manip_ex_5}).

\begin{figure}[!htbp]
\begin{center}
\begin{subfigure}[t]{0.32\linewidth}
\includegraphics[width=\linewidth, keepaspectratio]{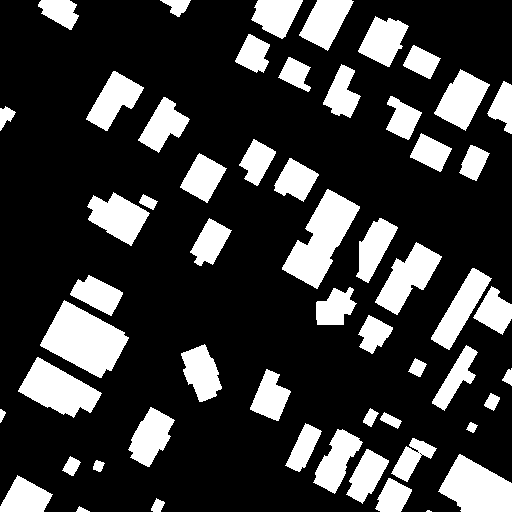}
\caption{Semantic Map}
\end{subfigure}
\begin{subfigure}[t]{0.32\linewidth}
\includegraphics[width=\linewidth, keepaspectratio]{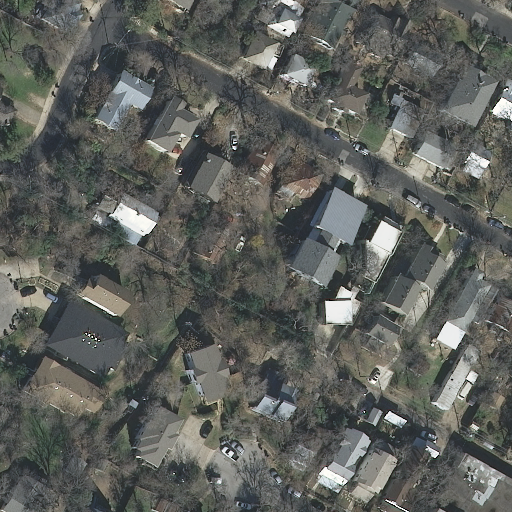}
\caption{Pristine Image}
\end{subfigure}
\begin{subfigure}[t]{0.32\linewidth}
\includegraphics[width=\linewidth, keepaspectratio]{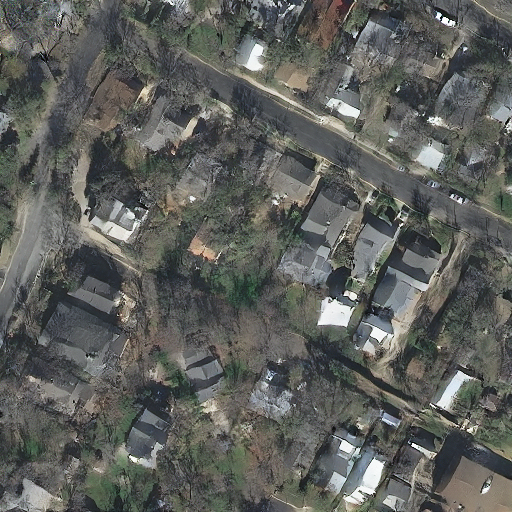}
\caption{Synthetic Image}
\end{subfigure}
\end{center}
\vspace{-0.7cm}
\caption{An example of imagel/label pair from Building2Sat dataset along with a pix2pixHD generated image.}
\label{fig:inria_ex}
\vspace{-0.3cm}
\end{figure}

\begin{figure}[!htbp]
\begin{center}
\begin{subfigure}[t]{0.45\linewidth}
\includegraphics[width=\linewidth, keepaspectratio]{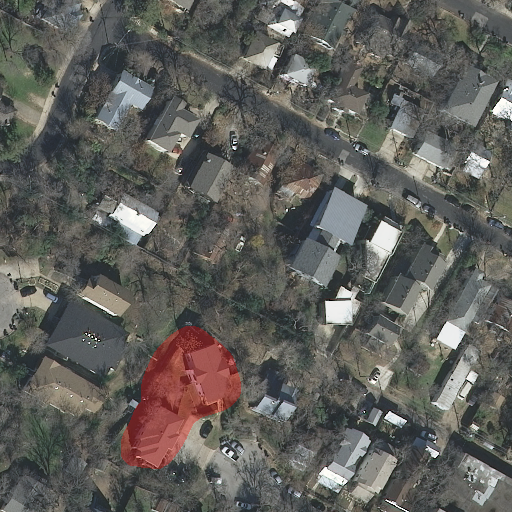}
\caption{Pristine Satellite Image, overlaid on removal mask}
\end{subfigure}
\hspace{0.1cm}
\begin{subfigure}[t]{0.45\linewidth}
\includegraphics[width=\linewidth, keepaspectratio]{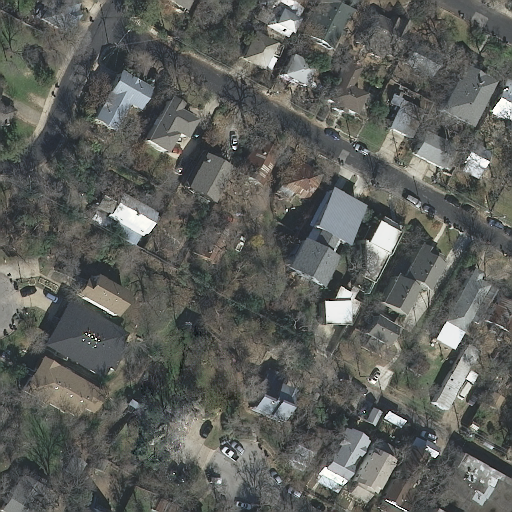}
\caption{Manipulated Image (post blending)}
\end{subfigure}
\end{center}
\vspace{-0.7cm}
\caption{An illustration of the proposed method on Building2Sat image, where a building has been removed.}
\label{fig:manip_ex_4}
\end{figure}

\begin{figure}[!htbp]
\begin{center}
\begin{subfigure}[t]{0.45\linewidth}
\includegraphics[width=\linewidth, keepaspectratio]{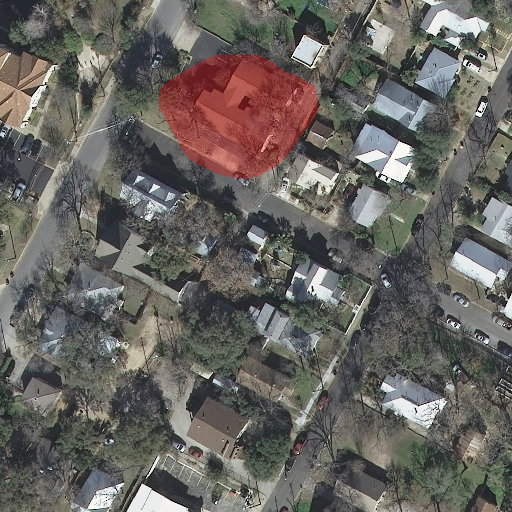}
\caption{Pristine Satellite Image, overlaid on removal mask}
\end{subfigure}
\hspace{0.1cm}
\begin{subfigure}[t]{0.45\linewidth}
\includegraphics[width=\linewidth, keepaspectratio]{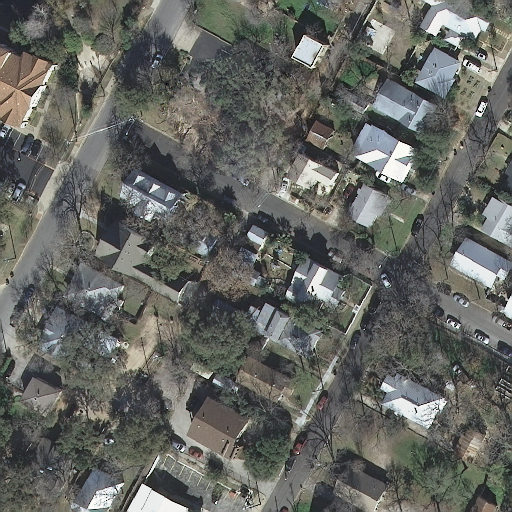}
\caption{Manipulated Image (post blending)}
\end{subfigure}
\end{center}
\vspace{-0.7cm}
\caption{An illustration of the proposed method on Building2Sat image, where a building has been removed.}
\label{fig:manip_ex_5}
\vspace{-0.3cm}
\end{figure}

\paragraph{Extensions to other datasets (Building2Sat and Cityscapes):}
The proposed framework above can be easily extended to semantic maps of different nature than those present in map2sat-urban dataset. Utilizing pix2pixHD as our GAN, we show that our method can be used to manipulate building segmentation maps to satellite images, trained using a dataset we call building2sat. This dataset is based on the Inria Aerial Image Labeling Dataset \cite{maggiori2017dataset}, which contains 180 5k x 5k image/label pairs from 5 locations: Austin, Chicago, Kitsap, Tyrol and Vienna. We select the 36 images from Austin to constrain the domain of the images, and split each 5k x 5k image into 500 x 500 tiles to obtain 3,600 image/label pairs. An example of an image/label pair along with a pix2pixHD generated image is shown in Figure~\ref{fig:inria_ex}, while Figures~\ref{fig:manip_ex_4}
and~\ref{fig:manip_ex_5} show examples of building removal using the proposed framework.

\begin{figure}[!htbp]
\begin{center}
\begin{subfigure}[t]{0.45\linewidth}
\includegraphics[width=\linewidth, keepaspectratio]{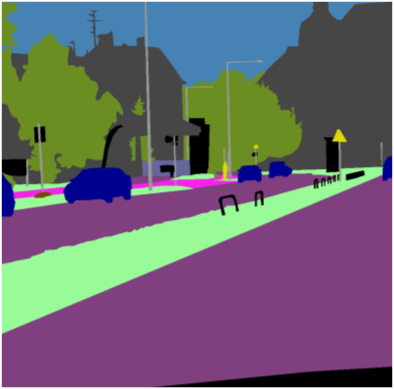}
\caption{Pristine Roadmap}
\end{subfigure}
\hspace{0.1cm}
\begin{subfigure}[t]{0.45\linewidth}
\includegraphics[width=\linewidth, keepaspectratio]{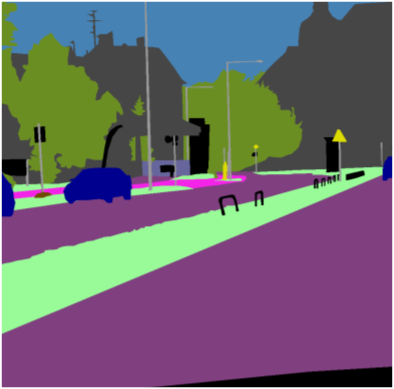}
\caption{Manipulated Roadmap}
\end{subfigure}
\\
\vspace{0.1cm}
\begin{subfigure}[t]{0.45\linewidth}
\includegraphics[width=\linewidth, keepaspectratio]{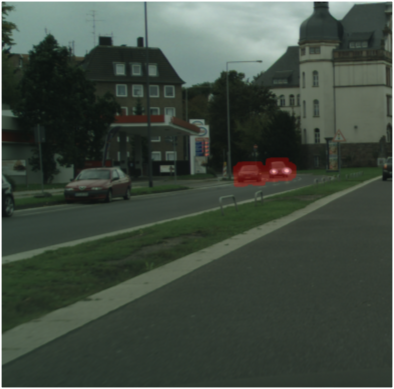}
\caption{Pristine Overhead Image, overlaid on removal mask}
\end{subfigure}
\hspace{0.1cm}
\begin{subfigure}[t]{0.45\linewidth}
\includegraphics[width=\linewidth, keepaspectratio]{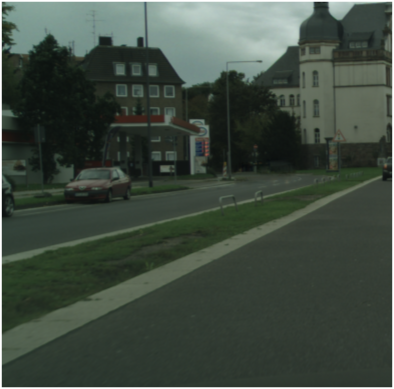}
\caption{Manipulated Image (post blending)}
\end{subfigure}
\end{center}
\vspace{-0.7cm}
\caption{Demonstration on image from cityscapes dataset, where a couple of vehicles have been removed.}
\label{fig:manip_ex_cityscapes_1}
\vspace{-0.5cm}
\end{figure}

The technique can be applied to any other dataset with images and semantic maps, even though it has only been demonstrated so far on aerial images. The Cityscapes dataset~\cite{cordts2015cityscapes}, which consists of 5000 images with city street scenes labeled with extremely rich semantic maps, is one such example where we used our method. We used a subset of this dataset that has high-quality annotations, that has 5000 image map pairs to retrain the Pix2PixHD GAN. Figure~\ref{fig:manip_ex_cityscapes_1} shows how our approach performed on this dataset with a few vehicles removed.

\begin{figure*}[!htbp]
    \centering
    \includegraphics[width=0.99\textwidth]{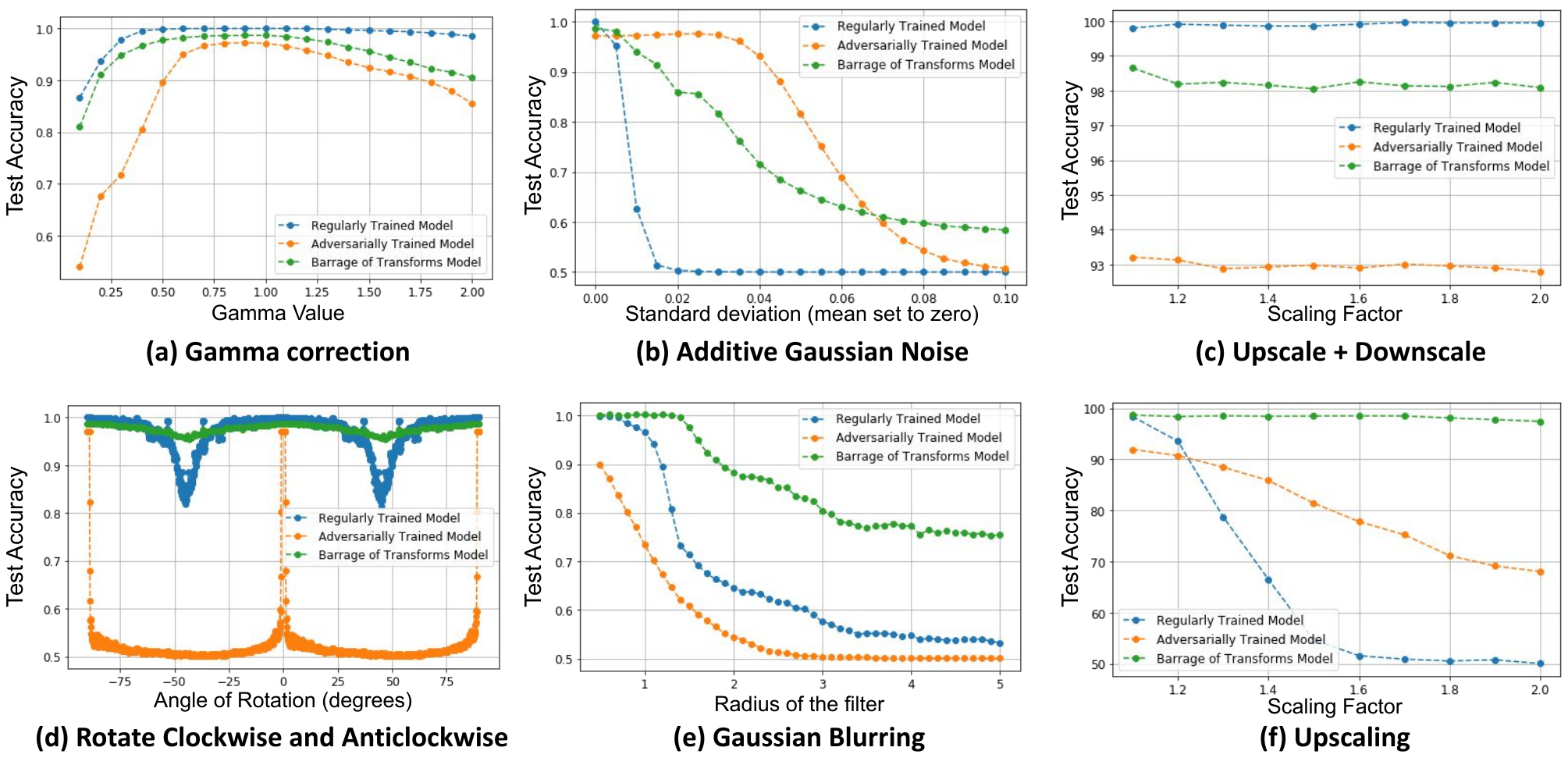}
    \vspace{-0.2cm}
    \caption{Deep learning based forensic techniques struggling to detect post-processed GAN generated images.}
    \label{fig:robustness_check}
\end{figure*}



\subsection{Quantitative Results}
Quantitative evaluation of the suggested technique is rather tricky due to the fact that the GAN-generated image is ultimately blended with the original image. Nevertheless, it is evident that the final blended image can only look decent if the GAN can memorize the map and image pairing and generate images with a highly similar appearance when the original map is input. To quantify this capacity of GANs, we relied on observing three standard metrics that are typically used in literature. (1) Fréchet Inception Distance (FID)~\cite{heusel2017gans}, (2) Kernel Inception Distance (KID)~\cite{Binkowski2018DemystifyingMG}, and (3) Structural Similarity Index (SSIM)~\cite{ssim_eg}. 
The following is how all three metrics are obtained. Excluding the manipulated map regions, pristine image and GAN-generated image from the manipulated map are divided into overlapping patches of size 64x64. These patches are used to evaluate the FID, KID, and SSIM scores. We used patches from approximately 20 examples from each dataset to calculate evaluation metrics. As the Build2Sat dataset has only masks for buildings, we can see that quantitative metrics are not that great on that dataset. Even though the quantitative metrics can be used for understanding the compatibility of GAN-generated images with our image manipulation framework, on a very high level, we urge the technique's end users rely more on manual observation than the quantitative numbers.




\vspace{-0.1cm}
\begin{table}[!hb]
\centering
\begin{tabular}{|c|c|c|c|}
\hline
\rowcolor[HTML]{C0C0C0} 
\textbf{Dataset} & \textbf{\begin{tabular}[c]{@{}c@{}}FID\\ ($\downarrow$)\end{tabular}} & \textbf{\begin{tabular}[c]{@{}c@{}}KID\\ ($\downarrow$)\end{tabular}} & \textbf{\begin{tabular}[c]{@{}c@{}}SSIM\\ ($\uparrow$)\end{tabular}} \\ \hline
Map2Sat-Urban & 40.77 & 0.027 & 0.65 \\ \hline
Building2Sat & 81.17 & 0.062 & 0.18 \\ \hline
CityScapes & 21.44 & 0.006 & 0.73 \\ \hline
\end{tabular}
\vspace{-0.2cm}
\caption{Quantitative metrics to measure the suitability of altered images. $\downarrow$ - Lower the better. $\uparrow$ - Higher the better.}
\end{table}





\section{Localizing Manipulated Regions}
\label{sec:exp_forensics}
\vspace{-0.2cm}

\begin{figure}[!htbp]
\begin{center}
\begin{subfigure}[t]{0.94\linewidth}
\includegraphics[width=\linewidth, keepaspectratio]{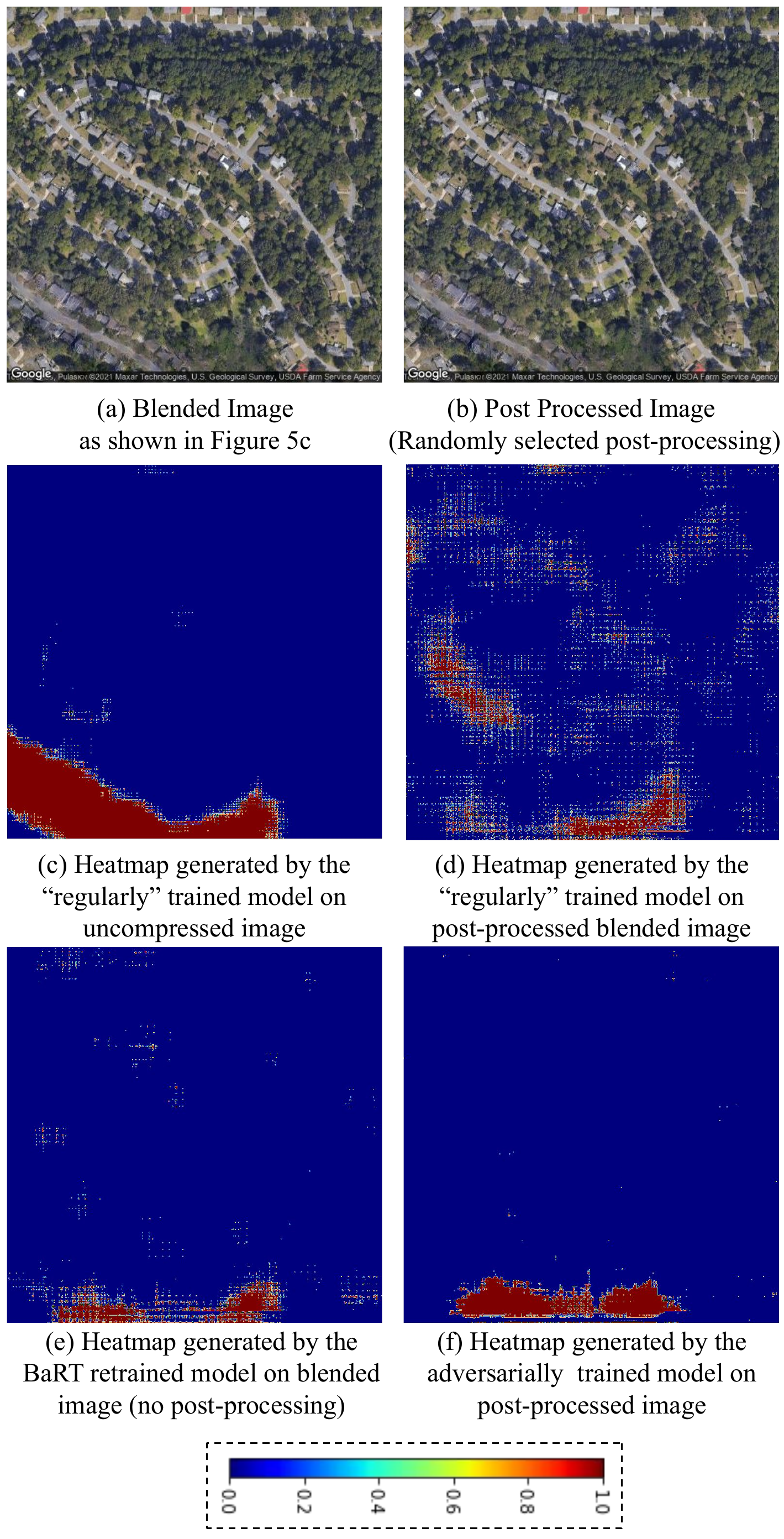}
\end{subfigure}

\end{center}
\vspace{-0.7cm}
\caption{Deep learning-based forensic techniques have trouble finding GAN-generated images that have been changed after they were produced.}
\vspace{-0.7cm}
\label{fig:BaRT_vs_RT}
\end{figure}

Given that trained GANs are capable of effectively forging images by removing/inserting/replacing objects, we undertook extensive experiments to determine if these manipulations may be detected by typical Convlutional Neural Network (CNN) based image forensic approaches.

The proposed framework for manipulation enables users to simply blend the GAN-generated image with the actual image. Thus, a substantial chunk of the altered image stays intact. This makes it more difficult to identify such modifications for image level manipulation detectors. However, it is possible to train a \textit{patch-based classification model}, which can be a convolutional neural network-based binary classifier trained on 64x64-pixel patches. We trained a comparable patch-based ResNet50 model for detecting and localizing GAN-generated or manipulated images. Using pristine and pix2pixHD GAN-generated images from the map2sat-urban dataset, we trained a patch-based model. Each image is divided into 64 x 64 non-overlapping patches in order to generate a dataset of 54,144 training images/patches and 6,016 validation images/patches (50 percent of them are GAN generated and the remaining half are pristine). When doing inference, we divide the input image into 64 x 64 patches and make predictions using our patch-based model with a stride of 1. With this setup, on the validation set, the trained CNN is able to achieve an AUC score of 99.99 percent and a maximum accuracy of 99.95 percent. But, we found that simple post-processing steps like rotation, scaling, gamma correction, or gaussian blurring can make the detector significantly less accurate, as shown by the blue curve in Figure~\ref{fig:robustness_check}.

\vspace{-0.1cm}
\textbf{Adversarial Training (AT):} To train the manipulation detectors that are robust to such post-processing steps, we conducted an experiment to determine if we could improve the robustness of the detectors by adversarially training~\cite{bunk2021adversarially} the model by attacking each mini-batch during the training process with an adversarial noise under L-infinity bound of ONE. The primary purpose of the experiment is to determine whether the adversarially trained model may provide increased robustness to post-processing operations in addition to its robustness against adversarial attacks. We discovered that the adversarially trained model provides robustness to multiple post-processing stages, but at the expense of a decrease in patch-level accuracy from 99 percent to nearly 85 percent on average. Heatmaps made by models that are only 85\% accurate tend to have a lot of noise, which makes them less reliable as real-time forensic detectors, as shown in Figure~\ref{fig:BaRT_vs_RT}.

\textbf{Barrage of Random Transform (BaRT)}-based re-training~\cite{raff2019barrage} is another experiment that has been conducted in an effort to strengthen the forensic method. In this experiment, images were randomly post-processed on the fly while the patch-based model was being trained. The experimental setup for training the detector is explained here. With a 50 percent probability, the sample undergoes post-processing. If a sample is selected for post processing, we select one of the following post processing steps: Gamma Correction (with different gamma values), Additive Gaussian Noise (by setting mean to zero and varying the standard-deviation), Gaussian Blurring (by varying the radius of the filter from 0.1 to 5.0), Upscaling, Upscaling + Downscaling (with different scaling factors), Rotate clockwise and anti-clockwise (with varying angles of rotation).

In summary, Barrage of Random Transform (BaRT)-based re-training enables an analyst to re-train by selecting common transforms/post-processing activities. In the majority of instances, the BaRT model outperforms the regular model and the AT model, as shown in Figure~\ref{fig:robustness_check}.

On the other hand, Adversarially trained (AT) model is “universal” and “blind” to any post-processing operations with its own limitations on L-infinity bound (maximum per-pixel perturbation). It is observed that AT model outperforms regular model in some cases but is not as good as BaRT model for most cases for which BaRT model was trained on. However, strength of the adversarially re-trained model arises from the fact that the model does not know beforehand what post-processing operations that might have taken place, but it can still detect most of the operations with more accuracy relative to regular models. 

AT and BaRT retraining have been proved to be more robust than naive CNN classifiers in a number of instances, however they have not yet reached accuracy levels that are regarded as reliable. However, the aforementioned forensic techniques cannot be guaranteed to be generalizable to other trained GANs, which is rather typical in our forgery pipeline given that users train GANs on smaller datasets each time they forge a new image. But, this opens the avenues for a combination of BaRT and AT models, which could prove to be far more robust.


\vspace{-0.3cm}
\section{Conclusion}
\label{sec:conclusion}
\vspace{-0.2cm}

This paper presents a framework for image manipulation with GANs. Specifically, we examined the fabrication of images from semantic maps utilizing two distinct architectures: CycleGAN and Pix2pixHD. While pix2pixHD generates images of great quality, CycleGAN has the advantage of training its generators to translate images in both directions. This advantage of CycleGAN enables manipulation of satellite images even in the absence of a semantic map, as the map can be built from the image itself using the same GAN. The methodology provided here preserves the vast majority of the original image's pixels, and can generate forgeries that are difficult for humans to identify visually. We also illustrated the ability of the proposed forging pipeline by demonstrating how it may circumvent many of the typical forensic techniques. As a future tasks, we are in the process of developing the similar image editing framework using other generative modeling techniques like stable diffusion. We are also exploring the possible ways to ensemble the BaRT and Adversarial training strategies to make image manipulation detectors more reliable.




{\small
\bibliographystyle{ieee_fullname}
\bibliography{egbib}
}

\end{document}